\pdfoutput=1
 \documentclass[10pt,twocolumn,twoside,journal]{IEEEtran}

\usepackage{cs}
\usepackage{mathsymb}
\usepackage{verbatim}
\usepackage{subfigure}
\usepackage{url}
\usepackage{booktabs}
\usepackage{cite}

\usepackage{graphicx}
\graphicspath{{./}{./Fig/}{./Figs/}{./photo_authors/}}

\usepackage{algorithm2e}
\let\savedalgorithm\algorithm
\let\savedendalgorithm\endalgorithm
\usepackage{algorithm}
\usepackage{amsmath}
\usepackage{url}
\usepackage{multirow}

\def\Integer{\mathbb{Z}^+}

\def\SQRTa{{\sc sqrt}\xspace}
\def\Hys{{\sc hys}\xspace}

\begin{document}

\title{Effective Pedestrian Detection  Using 
       Center-symmetric Local Binary/Trinary Patterns}

\author{
         Yongbin Zheng,
         Chunhua Shen,
         Richard Hartley,~\IEEEmembership{Fellow,~IEEE},
         and
         Xinsheng Huang
%\thanks
%{
%Copyright (c) 2010 IEEE. Personal use of this material is permitted.
%However, permission to use this material for any other purposes must be obtained
%from the IEEE by sending a request to pubs-permission@ieee.org.
%}
\thanks
{
Y. Zheng and X. Huang are with 
the National University of Defense Technology,
Changsha, Hunan, 410073, China.

This work was done when Y. Zhang visited NICTA Canberra Research
Laboratory and the Australian National University. 
}
\thanks
{
C. Shen and R. Hartley are with NICTA, Canberra Research Laboratory,
Canberra, ACT 2601, Australia,
and also with the Australian National University, Canberra,
ACT 0200, Australia
(e-mail: chunhua.shen@nicta.com.au; 
         richard.hartley@nicta.com.au).
C. Shen is the corresponding author.

NICTA is funded by the Australian Government as represented by
the Department of Broadband, Communications and the Digital
Economy and the Australian Research Council through the ICT
Center of Excellence program.

C. Shen's research was also supported in part by the Australian Research
Council through its Special Research Initiative in Bionic Vision Science and
Technology grant to Bionic Vision Australia.
}
%\thanks
% {
% Color versions of one or more of the figures in this paper are available online
% at http://ieeexplore.ieee.org.
% }
}

\markboth{September 2010}
{Zheng
\MakeLowercase{\textit{et al.}}: Effective Pedestrian Detection 
       Using Center-symmetric Local Binary/Trinary Patterns}

\maketitle

\begin{abstract}

Accurately detecting pedestrians in images plays a critically important role in many  computer
vision applications.  Extraction of effective features is the key to this task.  Promising features
should be discriminative, robust to various variations and easy to compute.  In this work, we
present novel features, termed dense center-symmetric local binary patterns (CS-LBP) and pyramid
center-symmetric local binary$/$ternary patterns (CS-LBP$/$LTP), for pedestrian detection. The
standard LBP proposed by Ojala et al.~\cite{c4} mainly captures the texture information. The
proposed CS-LBP feature, in contrast, captures  the gradient information and some texture
information. Moreover, the proposed dense CS-LBP and the pyramid CS-LBP$/$LTP are easy to implement
and computationally efficient, which is desirable for real-time applications.  Experiments on the
INRIA pedestrian dataset show that the dense CS-LBP feature with linear supporct vector machines
(SVMs) is comparable with the
histograms of oriented gradients (HOG) feature with linear SVMs, and the pyramid CS-LBP$/$LTP
features outperform both HOG features with linear SVMs and the start-of-the-art pyramid HOG (PHOG)
feature with the histogram intersection kernel SVMs. We also demonstrate that the
combination of our pyramid CS-LBP feature and the PHOG feature could significantly improve the
detection performance---producing state-of-the-art accuracy on the INRIA pedestrian dataset.

\end{abstract}


\section{Introduction}

  \IEEEPARstart{T}{he} ability to detect pedestrians in images has a major impact on applications such as video
surveillance~\cite{Surveillance}, smart vehicles~\cite{PedestrianVehic,PedestrianSmartCar},
robotics~\cite{PedestrianRobot}. Changing variations in human body poses and clothing, combined
with varying cluttered backgrounds and environmental conditions, make this problem far from being
solved. Recently, there has been a surge of interest in pedestrian detection
~\cite{Harr,Mikolajczyk-ECCV-2004,C1,HOGphdthesis,regioncovariant,ExperimentalPedestrianClassification,
covariancefeatures,dollarCVPR09peds,C2,MonocularPedestrian,EnzweilerPedestrianDetection,PedtrianDCrowedScenes,
DetectingShapeletFeatures,CascadeHistogramOrientied}.
One of the leading approaches for this problem is based on sequentially applying a classifier at all
the possible subwindows, which are obtained by exhaustively scanning the input image in different scales
and positions. For each sliding window, certain feature sets are extracted and fed to the
classifier, which is trained beforehand using a set of labeled training data of the same type of
features. The classifier then determines whether the sliding window contains a
pedestrian or not.

Driven by the development of object detection and classification, promising performance on
pedestrian detection have been achieved by:
\begin{enumerate}
\item
using discriminative and robust image features, such as
Haar wavelets~\cite{Harr},
region covariance~\cite{regioncovariant,covariancefeatures},
HOG~\cite{C1,HOGphdthesis} and
PHOG~\cite{Multi-levelHOG};
\item using a combination of multiple complementary features~\cite{C2,CombinationFeatureExtraction};
\item
including spatial information~\cite{Multi-levelHOG};
\item the choices of classifiers, such as support vector machines (SVMs) \cite{C1,Multi-levelHOG},
boosting \cite{Multi-CueOnboardPedestrianDetection,shen2010boosting}.
\end{enumerate}

Feature extraction is of the center importance here.
Features must be robust, discriminative, compact and
efficient. To date, HOG is still considered as
one of the state-of-the-art and most popular features used for pedestrian
detection~\cite{C1}. One of its drawbacks is the heavy computation.
Maji et al.~\cite{Multi-levelHOG} introduced the PHOG feature into pedestrian detection, and their experiments
showed that PHOG can yield better classification accuracy than the conventional
HOG and is much computationally simpler and have smaller dimensions. However, these HOG-like features, which capture
the edge or the local shape information, could perform poorly when the background is cluttered with noisy edges
\cite{C2}.

Our goal here is to develop a feature extraction method for pedestrian detection that, in comparison
to the state-of-the-art, is comparable in performance but  faster to compute. A conjecture is that,
if both the shape and texture information are  used as the features for pedestrian
detection, the detection accuracy is likely to increase. The center-symmetric local binary patterns
feature (CS-LBP)~\cite{c3}, which is a modified version of the LBP texture feature~\cite{c5}, inherits
the desirable properties of both texture features and gradient based features.
In addition, they are
computationally cheaper and easier to implement. Furthermore, CS-LBP can be extended to
center-symmetric Local Trinary Patterns (CS-LTP), which is more descriptive and less sensitive to
noise in uniform image regions.
In this work, we introduce the CS-LBP$/$LTP features into pedestrian detection:
\begin{enumerate}
\item
We propose the dense CS-LBP feature, in the approach similarity as the HOG feature~\cite{C1}, which was carefully developed to work well with linear SVMs for pedestrian detection.
\item
We propose the pyramid CS-LBP$/$LTP features, in the approach similarity as the PHOG feature~\cite{Multi-levelHOG}, which is muti-scale feature and producing the state-of-the-art accuracy with HIKSVMs on the INRIA pedestrian dataset.
\end{enumerate}
Experiments on the INRIA pedestrian dataset show that the dense CS-LBP feature with linear SVMs
performs as well as the HOG feature with linear SVMs, and the pyramid CS-LBP feature with
HIKSVMs~\cite{Multi-levelHOG} outperforms  the state-of-the-art PHOG features with HIKSVMs. The
pyramid CS-LTP feature can achieve even better performances.

 The key contributions of this work can be summarized as follows.
  \begin{enumerate}
    \item
    To our knowledge, it is the first time to apply the CS-LBP feature to pedestrian detection. The standard LBP feature captures the detailed texture information, which is usually harmful for pedestrian detection, e.g., the rich textures on the cloth of a pedestrian. Besides, the bin number of the standard LBP operator is 256, which leads a huge dimensional descriptor of a detection window. On the contrary, the CS-LBP feature captures the shape information and some salient texture information, which is very useful for pedestrian detection. The bin number of the CS-LBP is 16, which is much smaller than the standard LBP.

    \item
    We propose the CS-LTP feature, which is even more distinctive than the CS-LBP feature, for the first time.

    \item
    We apply the pyramid structure, which can can capture richer spatial information, to CS-LBP and CS-LTP for the first time.

    \item
    We show that the detection performance can be further improved significantly by combining our proposed pyramid CS-LBP$/$LTP features with the PHOG feature.

  \end{enumerate}

  The rest of the paper is organized as follows. In Section \ref{sec:Pre}, we
briefly describe the LBP operator, the LTP operator, and the CS-LBP operator. 
In Section \ref{sec:DenseApproach}, we give the details of the dense CS-LBP pedestrian detection
approach. In Section  \ref{sec:PLBPApproach}, 
we propose the pyramid CS-LBP$/$LTP
features based pedestrian detection approach. The results of numerous experiments and some study on
feature combination are presented in Section  \ref{sec:exp}. Section
\ref{sec:conclusion} concludes the paper.

\section{Preliminaries}
\label{sec:Pre}

\subsection{The LBP and LTP features}

LBP is a texture descriptor that codifies local primitives (such as curved edges, spots, flat
areas) into a feature histogram. LBP and its extensions outperform existing texture descriptors
both with respect to performance and to computational efficiency~\cite{c4}.

\begin{figure}[t]
         \begin{center}
            \includegraphics[width=0.45\textwidth]{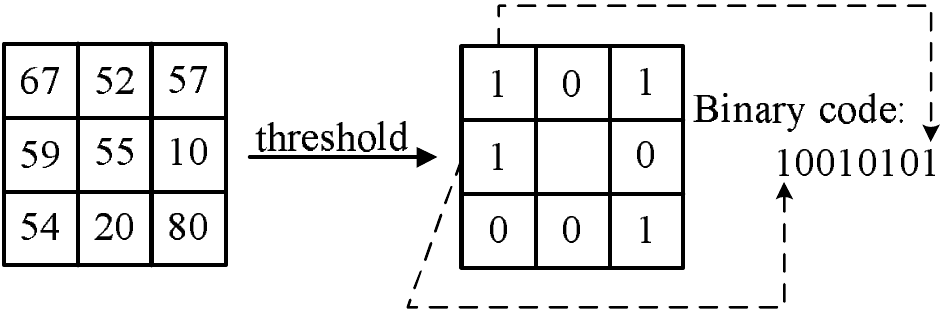}
         \end{center}
         \caption{Illustration of the basic LBP operator.}
         \label{fig:lbp}
       \end{figure}

The standard version of the LBP feature of a pixel is formed by thresholding the 3$\times$3-neighborhood
of each pixel with the center pixel's value . Let $g_c$ be the center pixel graylevel and
$g_i\,(i=0,1,\cdots,7)$ be the graylevel of each surrounding pixel. If $g_i$ is smaller than $g_c$,
the binary result of the pixel is set to 0, otherwise to 1. All the results are  combined to a
8-bit binary value. The decimal value of the binary is the LBP feature. See Fig.~\ref{fig:lbp} for an
illustration of computing the basic LBP feature.

\begin{figure}[t]
         \begin{center}
            \includegraphics[width=0.5\textwidth]{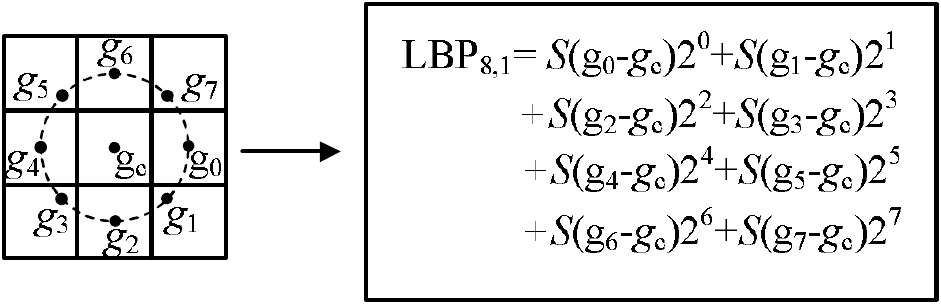}
         \end{center}
         \caption{The LBP operator of a pixel's circular neighborhoods with $r=1$, $p=8$.}
         \label{fig:lbp8}
     \end{figure}

In order to be able to cope with textures at different scales, the original LBP has been extended to
arbitrary circular neighborhoods~\cite{c5} by defining the neighborhood as a set of sampling points
evenly spaced on a circle centered at a pixel to be labeled. It allows any radius and number of
sampling points.  Bilinear interpolation is used when a sampling point does not fall in the center
of a pixel. Let LBP$_{p,r}$ denote the LBP feature of a pixel's  circular neighborhoods, where $r$
is the radius of the circle and  $p$ is the number of sampling points on the circle. The LBP$_{p,r}$
can be computed as follows:
\begin{equation}
\label{e1}
{\rm LBP}_{p,r}= \sum_{i=0}^{p-1}S(g_i-g_c)2^i, \,  S(x)=
\begin{cases}
1 & \text{if $x \geq 0$,}\\
0 & \text{ otherwise.}
\end{cases}
\end{equation}
Here $g_c$ is the center pixel's graylevel and $g_i \, (i=0,1,\cdots,7)$ is the graylevel of each
sampling pixel on the circle. See Fig.~\ref{fig:lbp8} for an illustration of computing the LBP feature of
a pixel's circular neighborhoods with $r=1$ and $p=8$.
Ojala {et al.}~\cite{c5} proposed the concept of ``uniform patterns" to reduce the number of possible LBP patterns while keeping its discrimination power.
An LBP pattern is called uniform if the binary pattern contains at
most two bitwise transitions from $0$ to $1$ or {\em vice versa}
when the bit pattern is considered circular.
For example, the bit pattern $11111111$ (no transition), $00001100$
(two transitions) are uniform
whereas the pattern $01010000$ (four transitions) is not.
The uniform pattern constraint reduces the number of LBP patterns
from $256$ to $58$ and is successfully applied to face detection
in~\cite{c6}.

\begin{figure}[t]
         \begin{center}
            \includegraphics[width=0.50\textwidth]{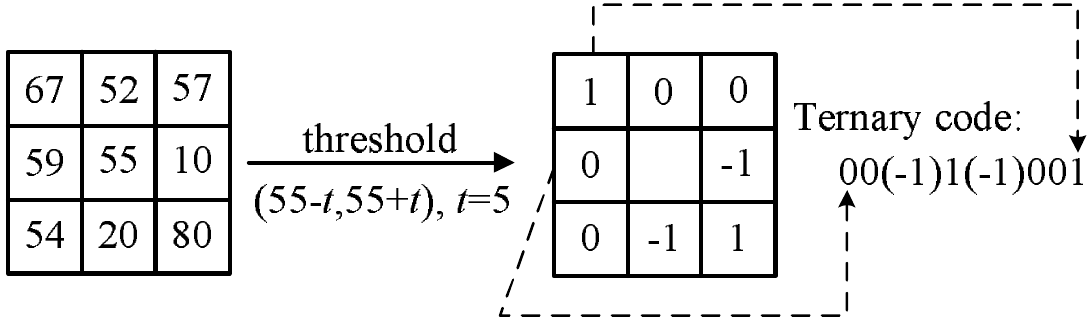}
         \end{center}
         \caption{Illumination of the basic LTP operator.}
         \label{fig:ltp}
       \end{figure}

\begin{figure}[t]
         \begin{center}
            \includegraphics[width=0.40\textwidth]{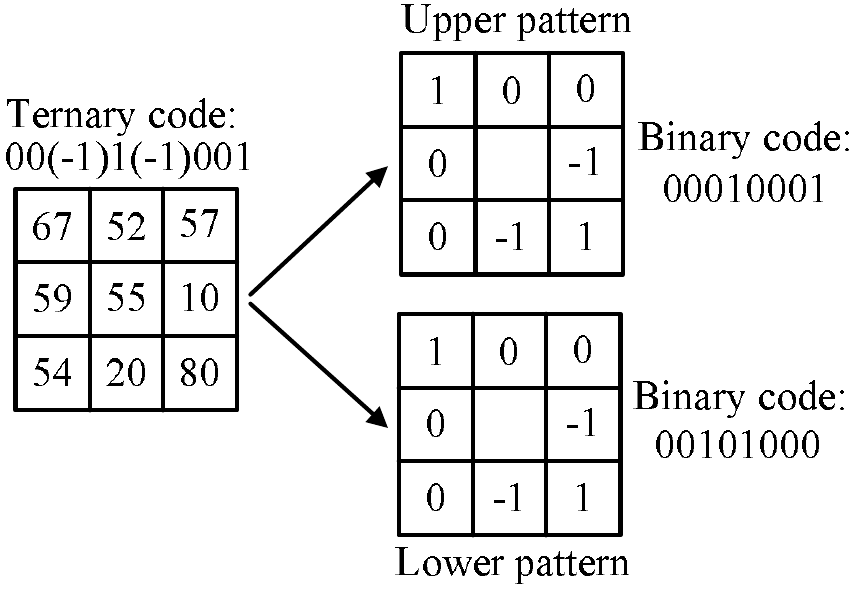}
         \end{center}
         \caption{Splitting the LTP code into positive and negative LBP codes.}
         \label{fig:ltp2}
     \end{figure}

In order to make LBP less sensitive to noise, particularly in near-uniform image regions, Tan and Triggs~\cite{LTP} extended LBP to 3-valued codes, called local trinary patterns (LTP). If each surrounding graylevel $g_i$ is  in a zone of width $\pm t$ around the
center graylevel $g_c$, the result value is quantized to 0. The value is quantized to $+1$ if $g_i$
is above this and is quantized to $-1$ if $g_i$ is below this. The LTP$_{p,r}$ can be computed as:
\begin{equation}
\label{eltp}
{\rm LTP}_{p,r}= \sum_{i=0}^{p-1}S(g_i-g_c)3^i, \,  S(x)=
\begin{cases}
   1 & \text{if $x \ge t$,}\\
   0 & \text{if $ |x| < t$,}\\
   -1 & \text{if $x \le t $,}
\end{cases}
\end{equation}
Here $t$ is a user-specified threshold. Fig. \ref{fig:ltp} shows the encoding procedure of
LTP. For simplicity, Tan and Triggs \cite{LTP} used a coding scheme that splits each ternary
pattern into its positive and negative halves as illustrated in Fig. \ref{fig:ltp2}, treating these
as two separate channels of LBP codings for which separate histograms are computed, combining the
results only at the end of the computation.

%-------------------------------------------------------------------------
\subsection{The CS-LBP/LTP patterns}

\begin{figure}[t]
      \begin{center}
         \includegraphics[width=0.45\textwidth]{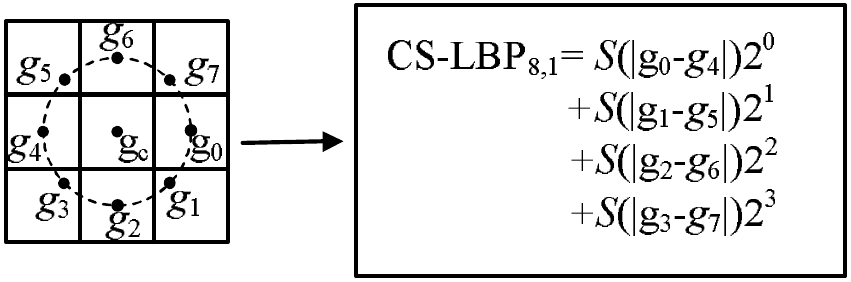}
      \end{center}
      \caption{The CS-LBP features for a neighborhood of $8$ pixels.}
      \label{cslbp}
   \end{figure}

The CS-LBP is another modified version of LBP. It is originally proposed to alleviate
some drawbacks of the standard LBP.
For example, the original LBP histogram could be very long and  the original
LBP feature is not robust on flat images.
As demonstrated in Fig.~\ref{cslbp}, instead of comparing the
graylevel of each pixel  with the center pixel, the center-symmetric pairs of pixels are compared.
The CS-LBP features can be computed by:
\begin{align}
    \label{e2}
    \textrm{CS-LBP}_{p,r,t} &= \sum_{i=0}^{N/2-1}S(|g_i-g_{i+(N/2)}|)2^i, \\
 S(x) &=
\begin{cases}
1 & \text{if $x \geq t$,}\\
0 & \text{ otherwise.}
\end{cases}
\end{align}

\begin{figure*}[t]
         \begin{center}
         %\finegap
         \subfigure[Original image.]{
            \label{fig:LbpImage:a}
            \includegraphics[width=0.4\textwidth]{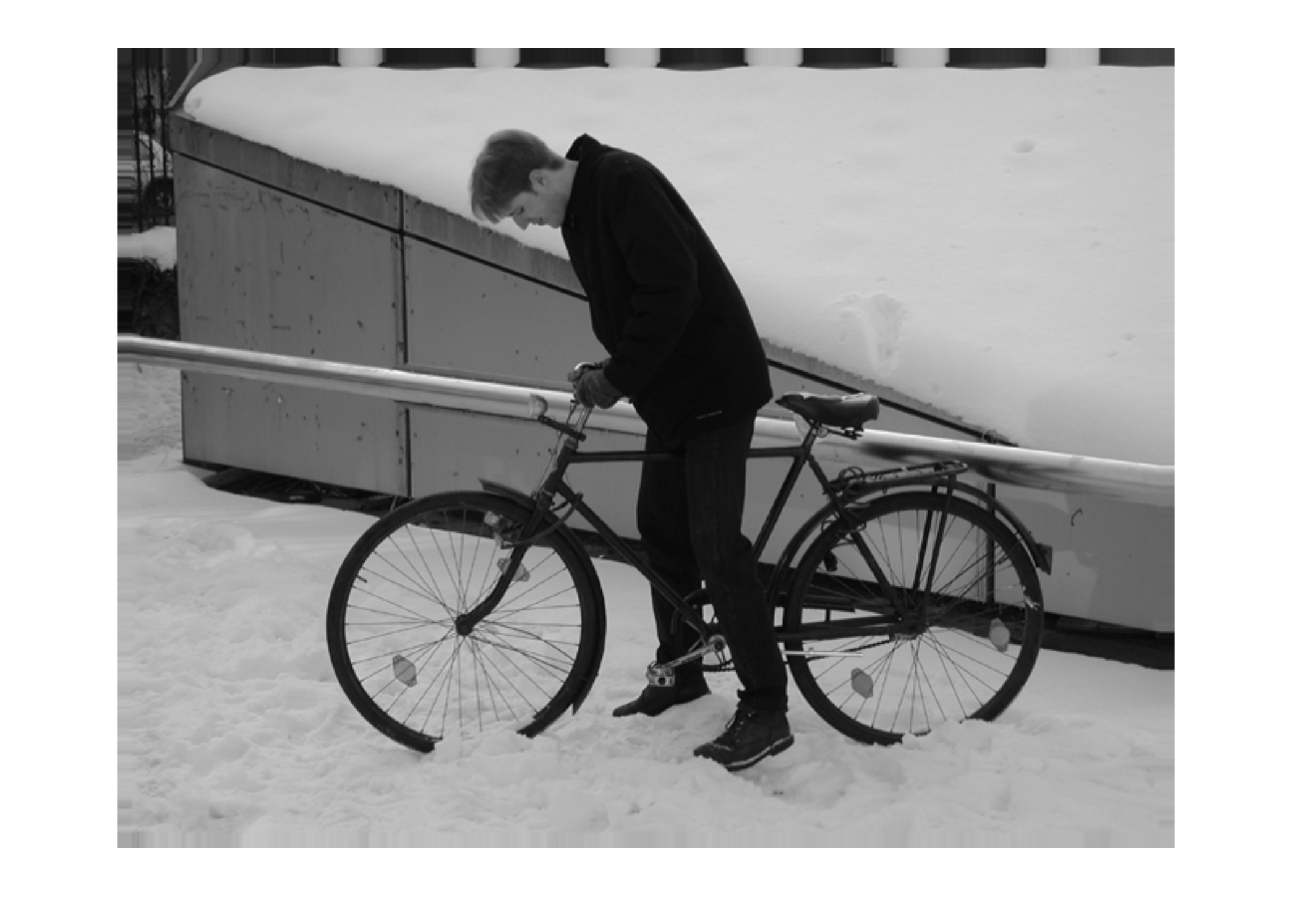} %
         }
       %  \hspace{.2in}
         %\finegap
         \subfigure[LBP image.]{
           \label{fig:LbpImage:b}
            \includegraphics[width=0.4\textwidth]{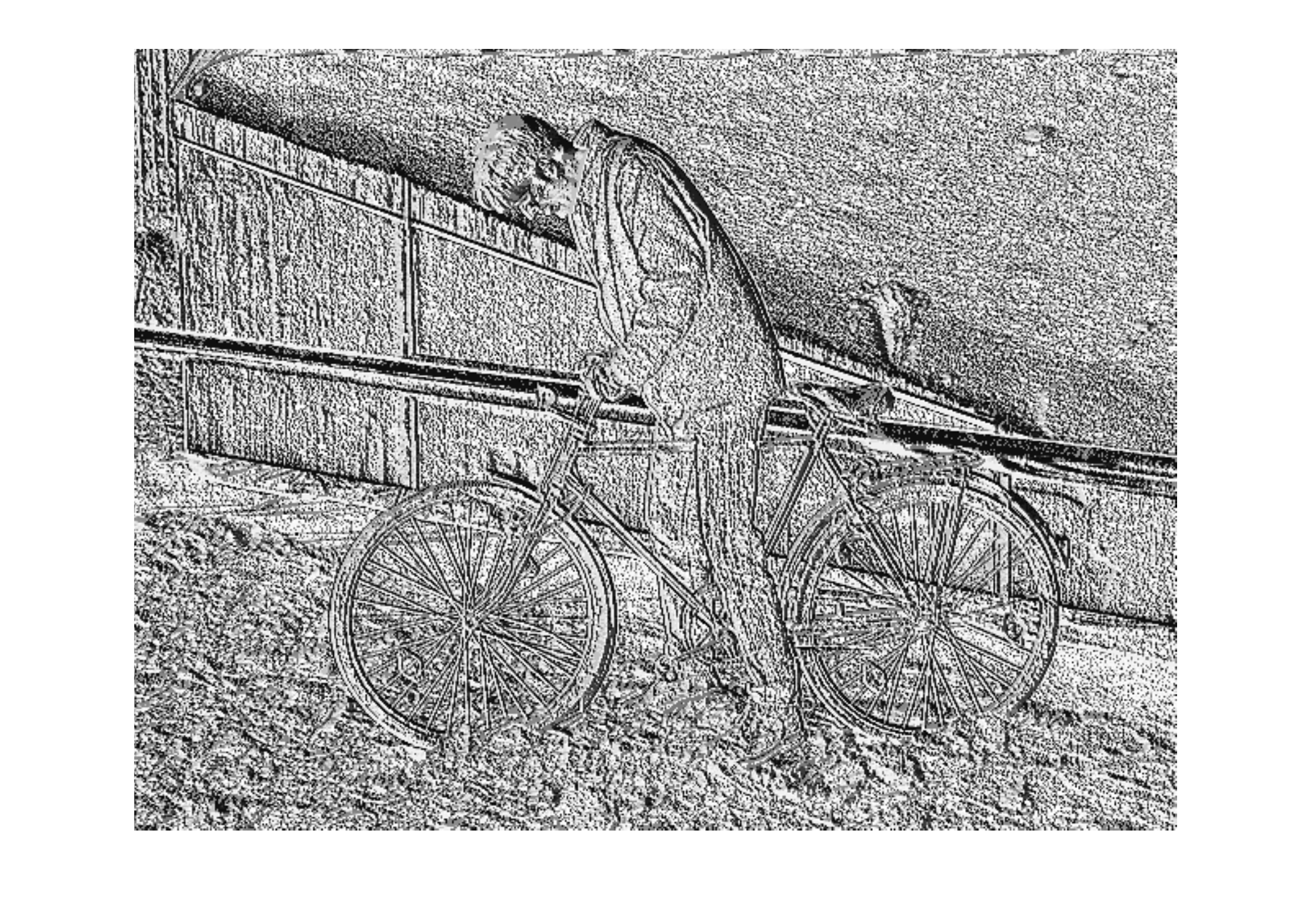}
         }
         \subfigure[Orientation bin image.]{
            \label{fig:LbpImage:c}
            \includegraphics[width=0.4\textwidth]{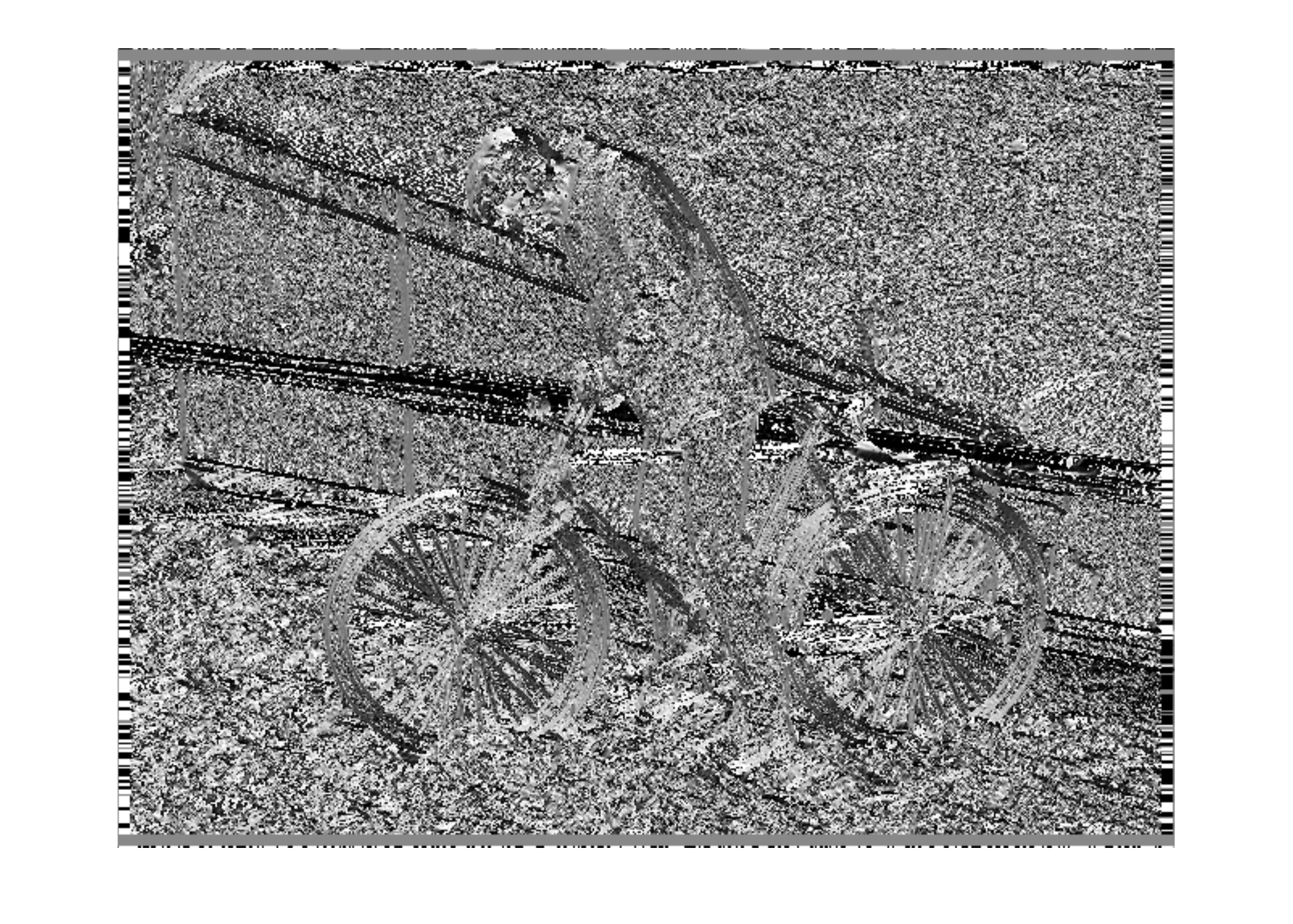} %
         }
     %    \hspace{.2in}
         %\finegap
         \subfigure[CS-LBP image.]{
           \label{fig:LbpImage:d}
            \includegraphics[width=0.4\textwidth]{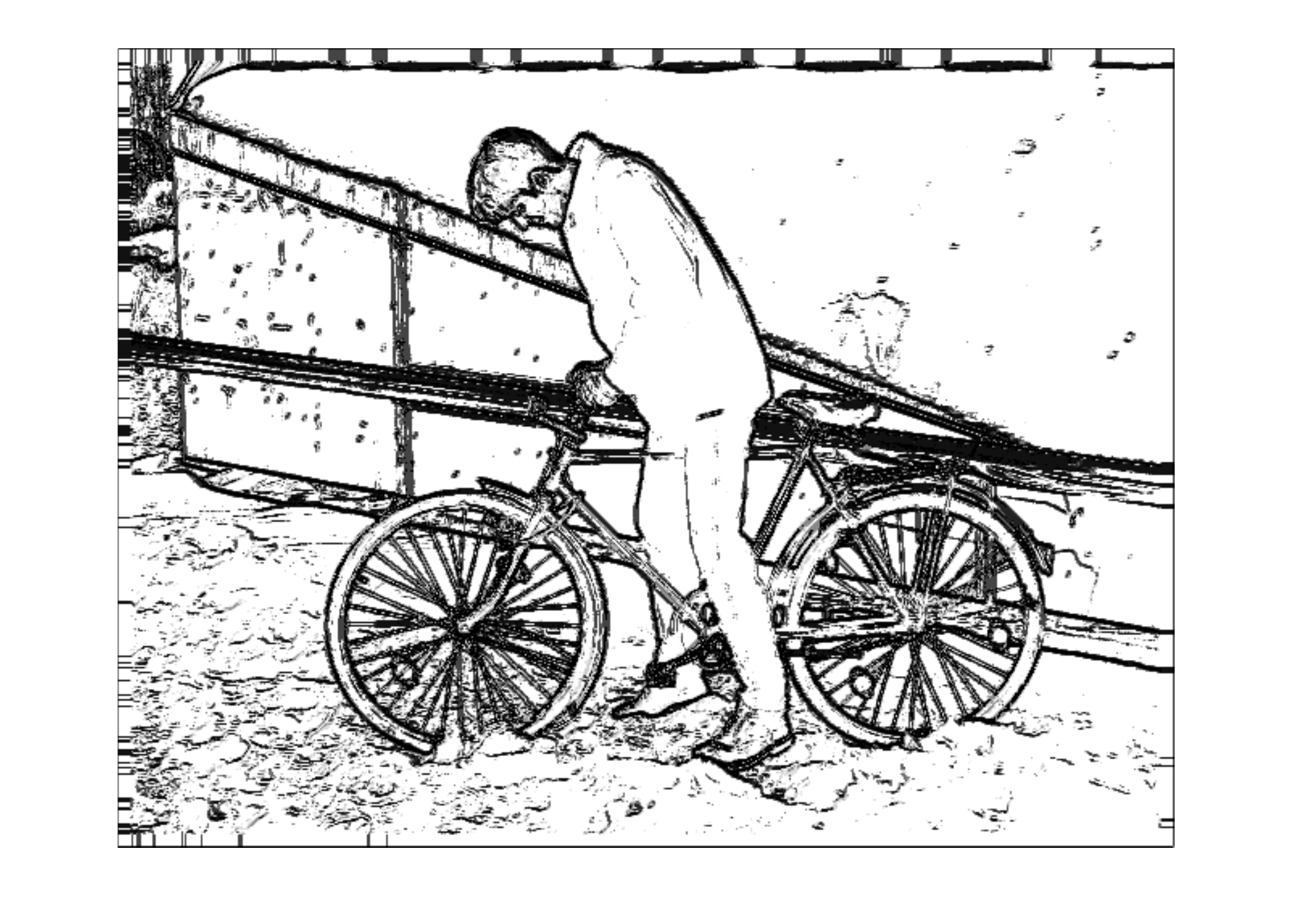}
         }
         \end{center}
         \caption{Example images of LBP, orientation bin and CS-LBP. (a) The original image selected from INRIA dataset. (b) The LBP image, which is obtained by replacing the graylevel of each pixel of the original image with the pixel's LBP value. (c) The orientation bin image, which is obtained by replacing the graylevel of each pixel of the original image by the pixel's orientation bin number. (d) The CS-LBP image, which is obtained by replacing the graylevel of each pixel of the original image by the pixel's CS-LBP value.}
         \label{FIG:LbpImage}
    \end{figure*}

Here $g_i$ and $g_{i+N/2}$ correspond to the graylevel of center-symmetric
pairs of pixels ($N$ in total) equally spaced on a circle of radius $r$.
Moreover, $t$ is a small value used to threshold the
graylevel difference so as to increase the robustness of the CS-LBP feature
on flat image regions. From the
computation of CS-LBP, we can see that the CS-LBP is closely related to the gradient operator,
because like some gradient operators, it considers graylevel differences between pairs of opposite pixels in
a neighborhood. In this way the CS-LBP feature takes advantage of
the properties of both the LBP and gradient based features. Fig.~\ref{FIG:LbpImage} shows images of
LBP, orientation bin and CS-LBP. The LBP image is obtained by replacing the graylevel of each pixel
of the original image with the pixel's LBP value; the orientation bin image is obtained by replacing
the graylevel of each pixel with its orientation bin number (the 16 orientation bins are evenly
spaced over $0^\circ-360^\circ$); the CS-LBP image is obtained by replacing the graylevel of each
pixel of the original image with the pixel's CS-LBP value. We can see that the CS-LBP captures the
edges and the salient textures. In~\cite{c3}, the authors used the CS-LBP descriptor to describe the
region around an interest point and their experiments show that the performance is almost equally
promising as the popular SIFT descriptor~\cite{SIFT}. The authors also compared the computational
complexity of the CS-LBP descriptor with the SIFT descriptor and it has been shown that the CS-LBP
descriptor is on average $2$ to $3$ times faster than the SIFT. That is because the CS-LBP feature
needs only simple arithmetic operations while the SIFT requires time consuming inverse tangent
computation when computing the gradient orientation.

Similarly as ``uniform LBP patterns", we propose ``uniform CS-LBP patterns" to reduce the original
CS-LBP pattern numbers. The possibility of each CS-LBP pattern is not equally distributed. 
The 8 patterns with bigger possibilities are called uniform while the rests are called non-uniform.
We computed the CS-LBP patterns of 741 images in the INRIA dataset (288 images containing
pedestrians and 453 images without pedestrians) with $t=0.022$ and found that 87.39\% of the
patterns are uniform, shown in Table~\ref{table:lbp}.

The CS-LTP patterns and the uniform CS-LTP patterns can be developed similarity as the CS-LBP and the uniform
CS-LBP.

\begin{table*}[tbh]
\begin{center}
\caption{The distribution of the CS-LBP patterns (uniform and non-uniform) with $t=0.022$
on the INRIA pedestrian dataset.}
\label{table:lbp}
\begin{tabular}{l | ccccccccc}
\hline
\hline
Uniform pattern  & 0000 & 0001 & 0011 & 0100 & 0111& 1000& 1101 & 1111 & Total \\
Percent. (\%)     & 7.67 & 7.34 & 2.19 & 5.65 & 3.47 & 2.28 & 3.52 & 55.26 &  87.39  \\
\hline
Non-uniform pattern & 0010 & 0101 & 0110 & 1001 & 1010 & 1011 & 1100 & 1110 & Total \\
Percent. (\%)       & 2.16 & 1.09 & 1.84 & 2.18 & 0.52 & 1.51 & 1.85 & 1.45 &  12.61 \\
\hline
\hline
\end{tabular}
\end{center}
\end{table*}
%\setlength{\tabcolsep}{0.8pt}

%-------------------------------------------------------------------------
\section{Pedestrian detection using dense CS-LBP feature}
\label{sec:DenseApproach}

\subsection{The approach }
\begin{figure*}[t]
      \begin{center}
         \includegraphics[width=0.62\textwidth]{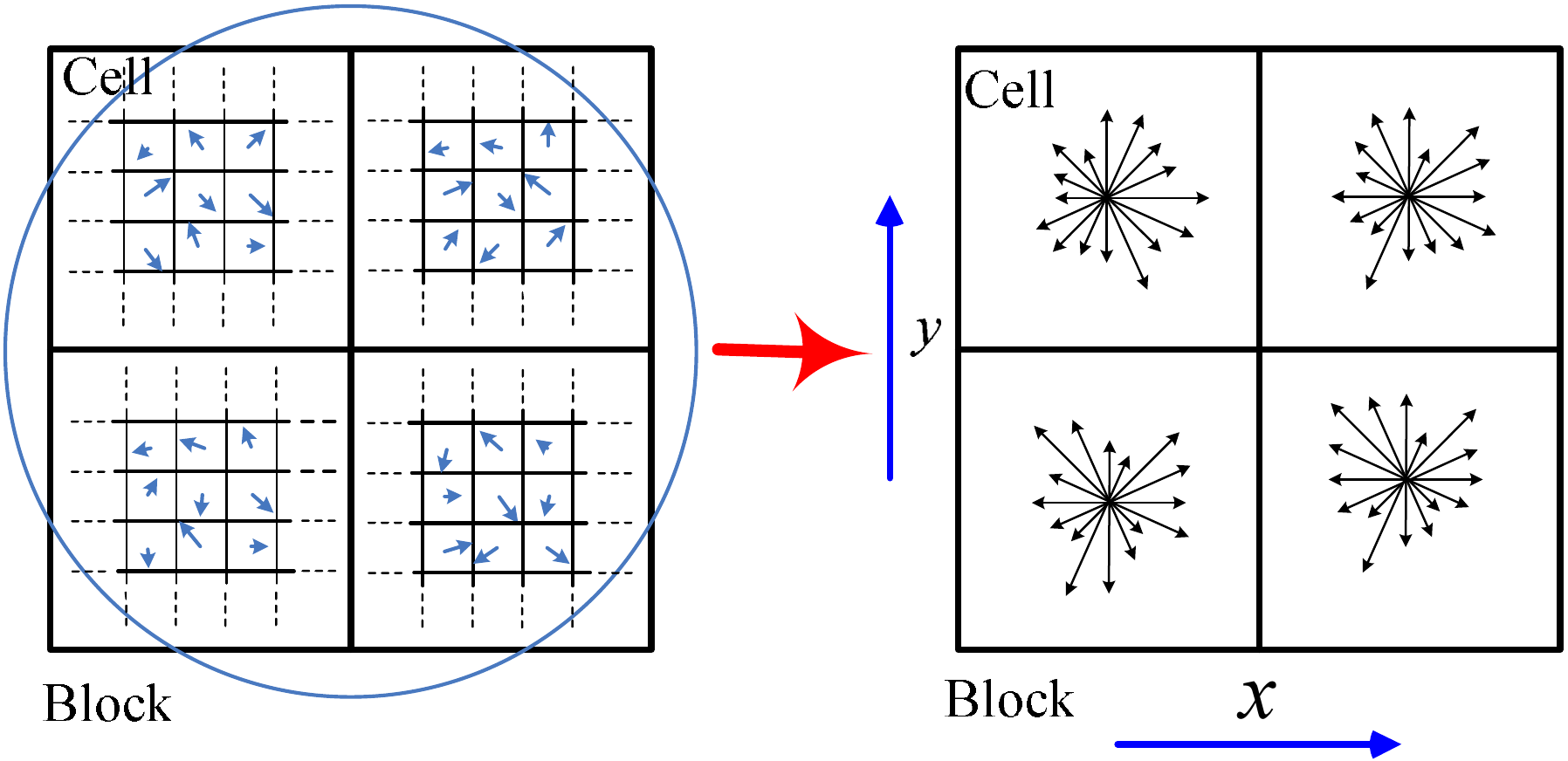}
      \end{center}
      \caption{A 3D histogram of each cell's locations ($x$ and $y$) and the cell's CS-LBP values (16 bins) is proposed for the block: The gradient magnitude and the CS-LBP value at each pixel in every cell are computed. The magnitudes are weighted by a Gaussian window centered in the middle of the block with $\sigma = 0.5 \times \rm blockwidth$, indicated by overlaid circle. The weighted values of all the points in a cell are accumulated into histograms by summarizing the contents over the cell. On the right of the figure, it shows 16 bins of each cell's histogram, with the length of each arrow corresponding to the magnitude of the histogram entry. }
      \label{CellBlock}
   \end{figure*}

In this section, we introduce the  implementation details of our dense CS-LBP feature
based pedestrian detection approach. Detailed comparisons between different parameter choices are carried out later. The key steps are as follows.
\begin{enumerate}
\item
We normalize the graylevel of the input image to reduce the illumination variance in different
images. After the graylevel normalization is performed, all input images have graylevel
ranging from 0 to 1.
\item
Each detection window is split into equally sized cells and the cells are grouped into bigger
blocks. The size of our detection window is $64 \times 128$ and the size of each block is $32 \times 32$ and each block contains $2 \times 2$ cells of $16 \times 16$ pixels, as shown in Fig.~\ref{CellBlock}. As in~\cite{C1}, there are overlaps among adjacent blocks (overlapping $1/2$ block).

\item
The 3D histogram of each block is computed similarly as the SIFT descriptor: The gradient magnitude and the CS-LBP value at each pixel in every cell are computed, as the arrows shown on the left of Fig.~\ref{CellBlock}. These are weighted by a Gaussian window centered in the middle of the block with $\sigma = 0.5 \times \rm blockwidth$, indicated by overlaid circle. The weighted values of all the points in a cell are accumulated into histograms by summarizing the contents over the cell. On the right of Fig.~\ref{CellBlock}, it shows 16 bins for the histogram of each cell, with the length of each arrow corresponding to the magnitude of the histogram entry. A 3D histogram of the cells' locations ($x$ and $y$ shown on the right of Fig.~\ref{CellBlock}) and the cells' CS-LBP values is proposed for the block. In order to avoid boundary effects in which the 3D histogram abruptly changes as a feature shifts from one cell to another, bilinear interpolation over horizontal and vertical dimensions is used to share the weights of the features between four nearest cells. Interpolation over CS-LBP value dimension is not carried out because the CS-LBP feature is quantized by its nature~\cite{c3}.

\item
The 3D histogram of each block is converted into a vector and is normalized.
% The commonly used
% normalization schemes are $\ell_1$-norm, $\ell_1$-\SQRTa, $\ell_2$-norm,
% $\ell_2$-\Hys as described in~\cite{C1}.
Let $v$ be the unnormalized descriptor, $\|v\|_k$ be its $k$-norm for $k=1,2$, and $\epsilon$ be a
small constant.
The commonly used normalization schemes are~\cite{C1}: 
\begin{enumerate}
    \item
$\ell_1$-norm,
$v\leftarrow v/(\|v\|_1 + \epsilon)$; 
    \item
    $\ell_1$-\SQRTa, $\ell_1$-norm followed by square root
$v\leftarrow v/\sqrt{(\|v\|_1 + \epsilon)}$; 
\item
    $\ell_2$-norm, $v\leftarrow v/\sqrt{(\|v\|^2_2 +
\epsilon)}$; 
\item
    $\ell_2$-\Hys, $\ell_2$-norm followed by clipping (limiting the maximum values
of $v$ to $0.2$) and re-normalizing. 
 \end{enumerate}
In our implementation, $\ell_1$-\SQRTa normalization gives the best result.
The difference between these normalization schemes
are not significant.
\item
The histograms of all the blocks in a detection window are concatenated to form a CS-LBP descriptor. This is used as
the input for the linear SVMs classifier.
\item
The detection window slides on the input images in all positions and scales, with a fixed scale factor 1.09 and a fixed step size $8\times 8$. The descriptor of each detection window is classified by the pretrained linear SVM classifier. As in \cite{HOGphdthesis}, non maximal suppression~\cite{meanshift2002} clustering is used to merge the multiple overlapping detections in the 3D position and  scale space.
\end{enumerate}

\subsection{Parameters selection}

There are various parameter configurations that can be chosen to optimize the performance of the CS-LBP feature based
detection approach. These include choosing the block size and cell size, $\sigma$ of the Gaussian
weighing window, using interpolate bilinearly over $x$ and $y$ dimensions when building the
histogram, the normalization method and the overlapping size of blocks.

We train a linear SVMs classifier using the training set described in Section \ref{sec:exp:setup} and
use the $1,132$ cropped human samples with size $70\times134$ (a margin of 3 pixels around each
side) from the test dataset as the positive test set. We randomly select $4,530$ patches with size
$64\times128$ from the 453 human free images in the test dataset as negative test set. Then we use
the pretrained classifier to classify between the positive samples and the negative samples. The
classification rate of the positive samples versus false positive rate is used to evaluate the
performances of different parameter selections.

\begin{figure*}[t!]
         %\centering
         \subfigure[]{
            \label{fig:Dcslbp:a}
            \includegraphics[width=0.47\textwidth]{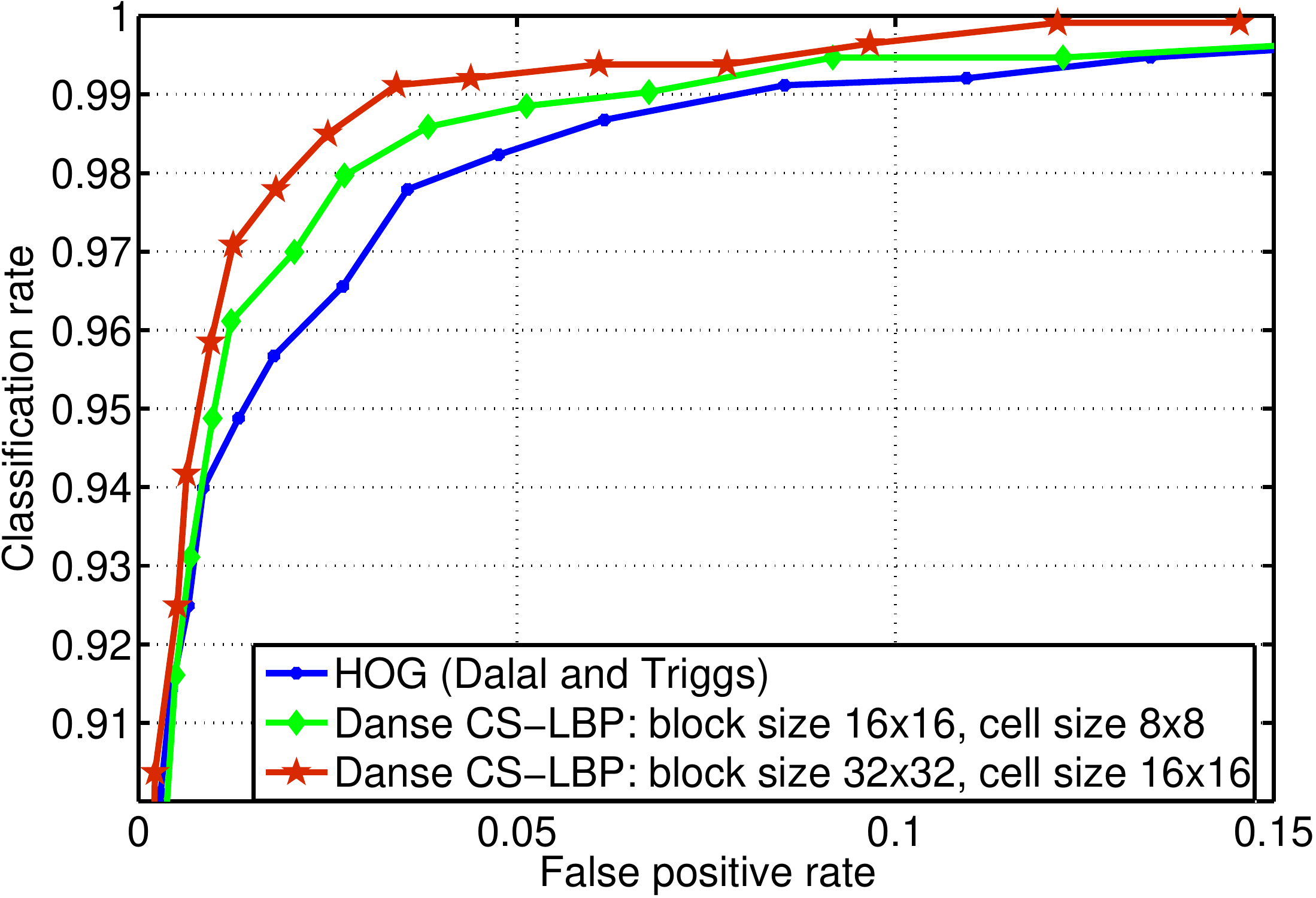}
         }
         \quad
         \subfigure[]{
            \label{fig:Dcslbp:b}
            \includegraphics[width=0.47\textwidth]{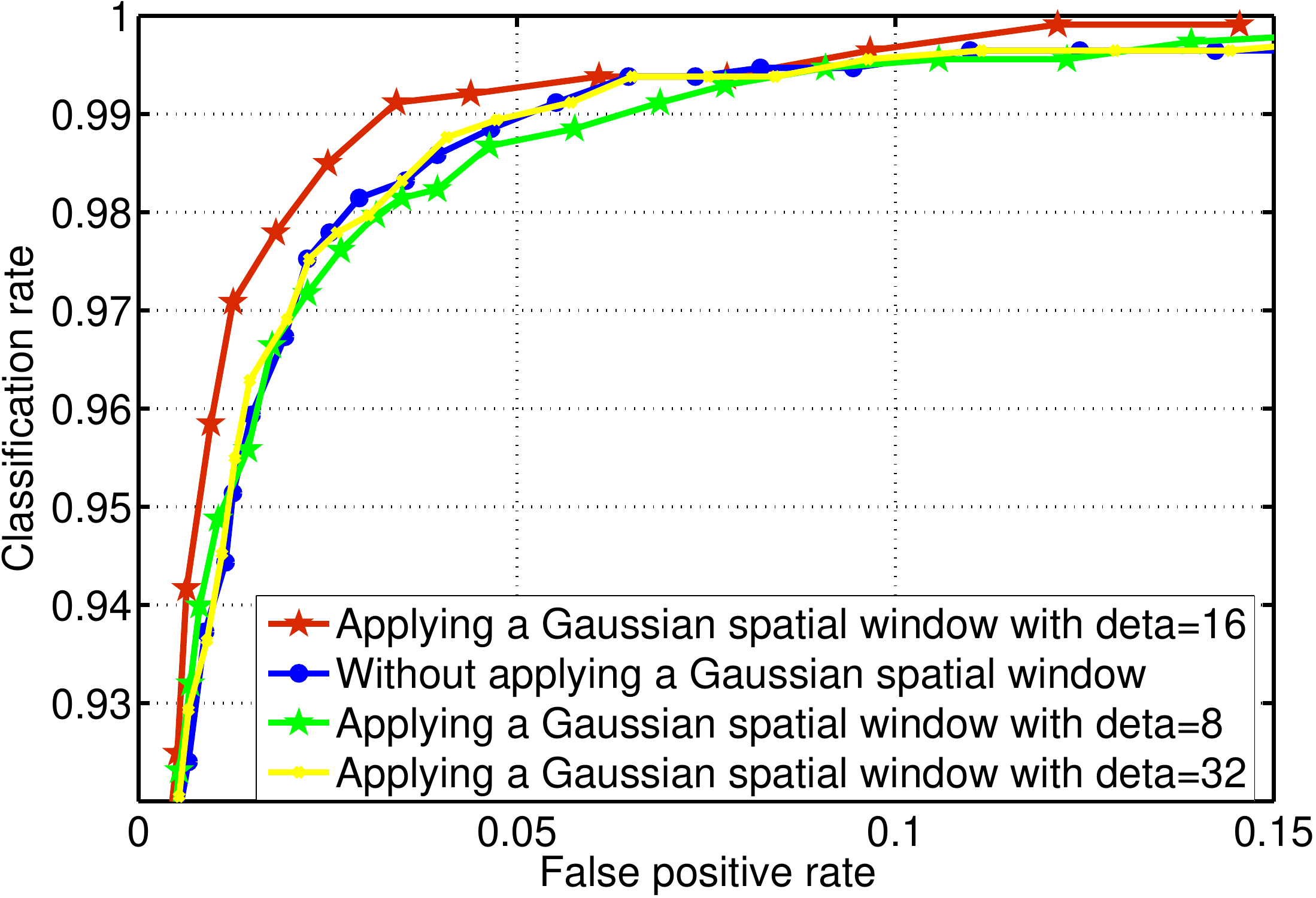}
         }
         \subfigure[]{
            \label{fig:Dcslbp:c}
            \includegraphics[width=0.47\textwidth]{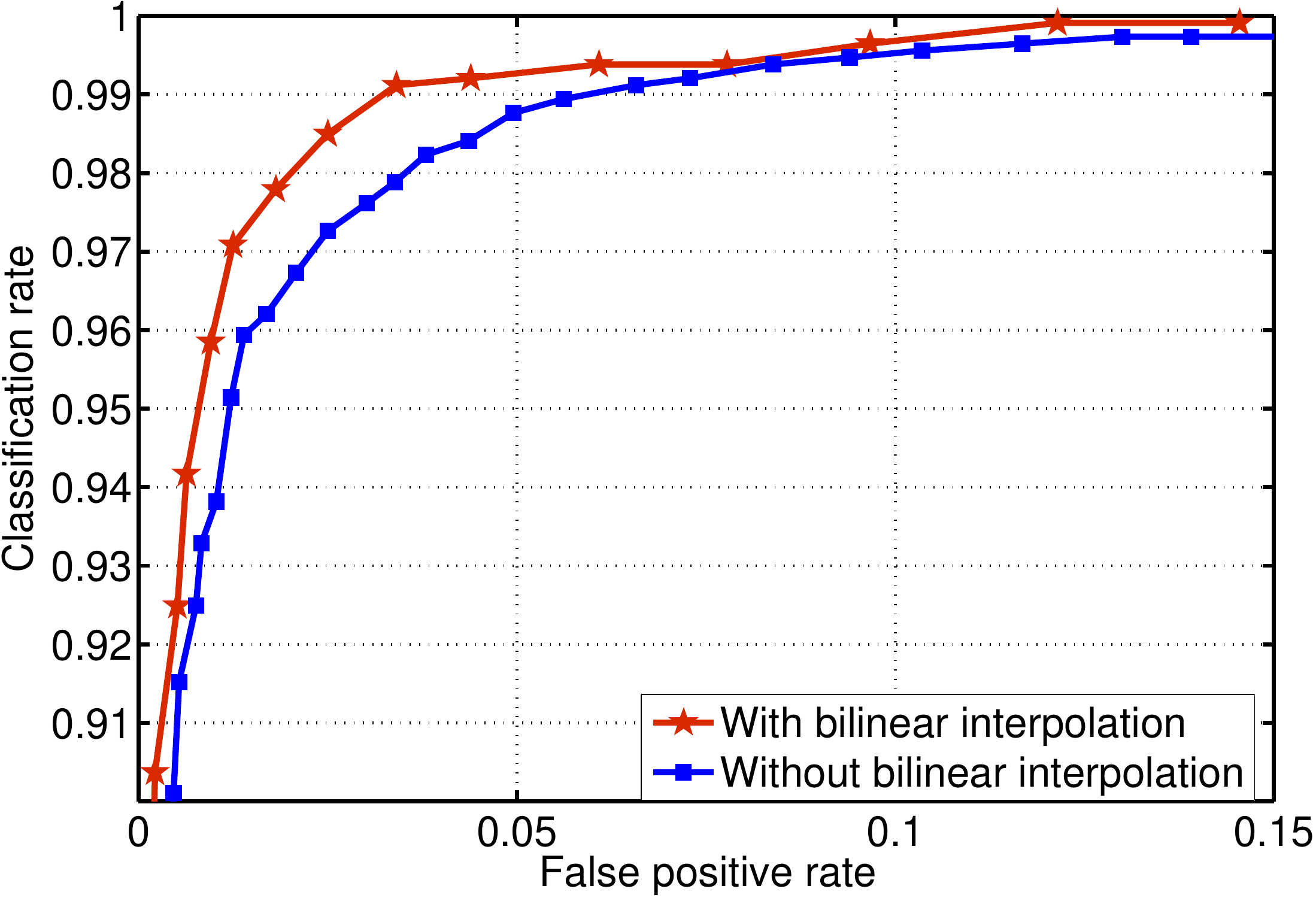}
         }
         \subfigure[]{
            \label{fig:Dcslbp:d}
            \includegraphics[width=0.47\textwidth]{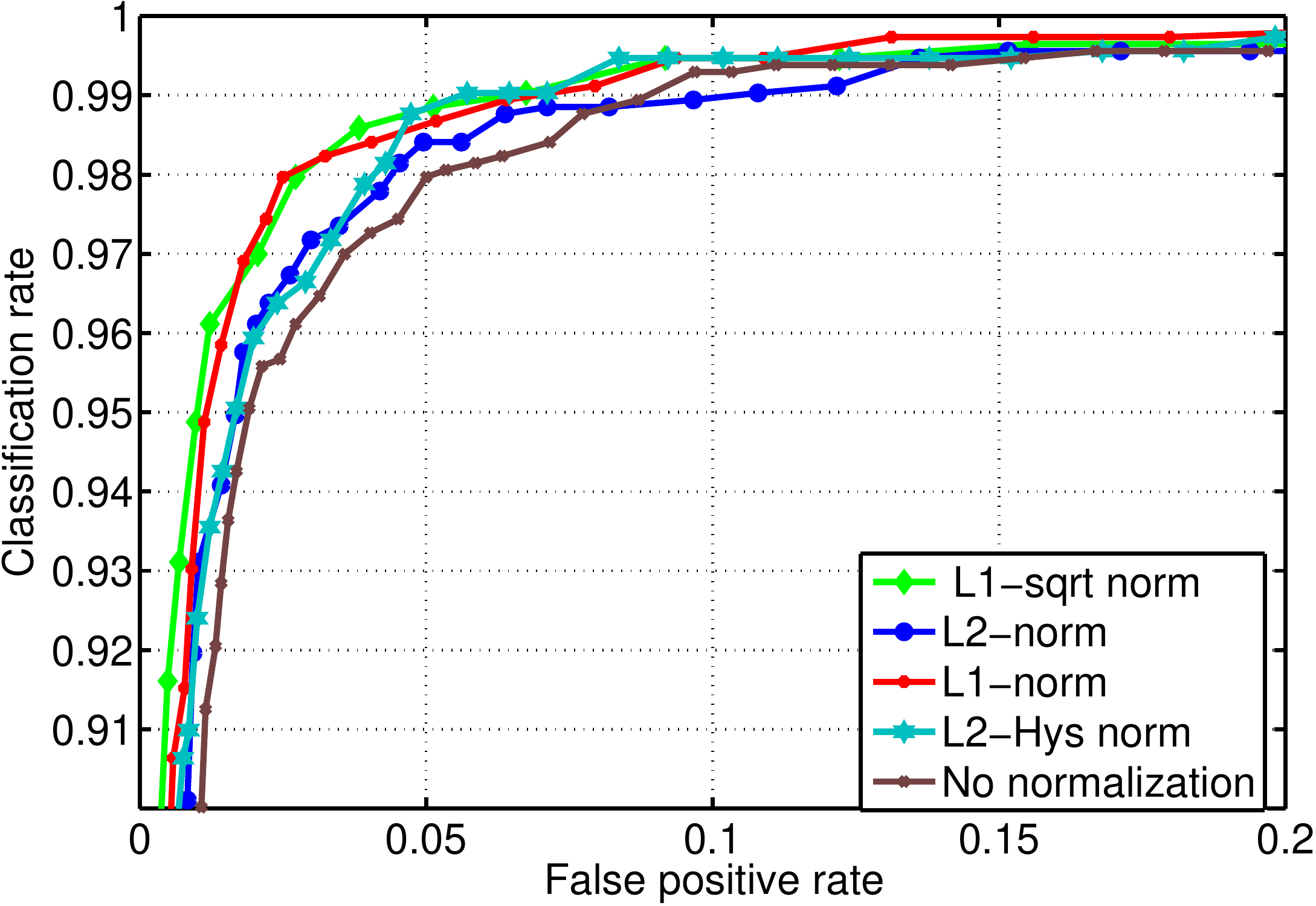}
         }
         \subfigure[]{
            \label{fig:Dcslbp:e}
            \includegraphics[width=0.47\textwidth]{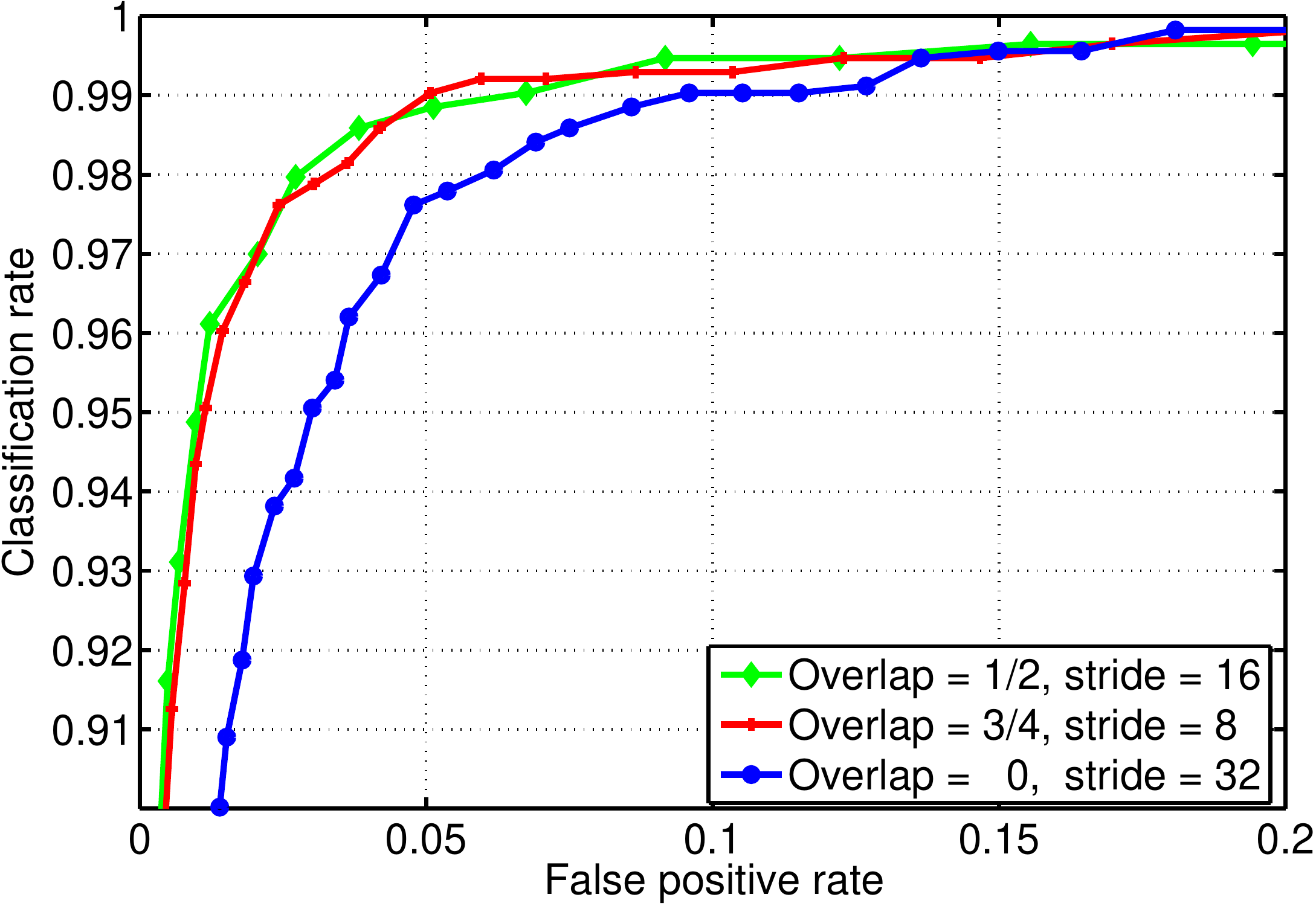}
         }
 \caption{Experimental results. 
         (a) Performance comparison of the CS-LBP feature with different block sizes
         and cell sizes.
         (b) Performance comparison of the CS-LBP feature with different Gaussian weight factor
         $\sigma$. (c) Performance comparison of the CS-LBP feature with and without bilinear
         interpolation. (d) Performance comparison of the CS-LBP features with different
         normalization methods. (e) Performance comparison of the CS-LBP features with different
         rate of overlapping.}
 \label{fig:Dcslbp}
 \end{figure*}

We compare the performances of our CS-LBP features with different block size and cell size configurations in
Fig.~\ref{fig:Dcslbp:a}. It shows that 32$\times$32 pixels blocks with 16$\times$16 pixels cells performs
better than 16$\times$16 pixels blocks with 8$\times$8 pixels cells.

We explore the effect of the Gaussian weight window in Fig.~\ref{fig:Dcslbp:b}. The results show that
a Gaussian weight window with $\sigma = 16$ (half block width) can improve the performance
significantly. However, if $\sigma $ is too big or small, the performance is almost identical as the case
when there is no Gaussian weight.

Fig.~\ref{fig:Dcslbp:c} shows that using bilinear interpolation when building the histogram of each
block can increase the performance.

We also evaluate four different normalization schemes in Fig.~\ref{fig:Dcslbp:d}. The schemes are: $\ell_2$-norm,
$\ell_2$-\Hys, $\ell_1$-norm, $\ell_1$-\SQRTa.
Fig.~\ref{fig:Dcslbp:d} shows that $\ell_1$-\SQRTa performs best and $\ell_1$-norm
performs very close to $\ell_1$-\SQRTa. $\ell_2$-\Hys and $\ell_2$-norm
are about $2\% $ worse than $\ell_1$-\SQRTa when false positive rate is $0.03$. The performance of without normalization is worst.

Fig.~\ref{fig:Dcslbp:e} shows the performance of overlapping blocks. We can see from
Fig.~\ref{fig:Dcslbp:e} that the detection rate increases when overlapping $1/2$ blocks, and overlapping
$3/4$ blocks performs equally to overlapping $1/2$. Overlapping $1/2$ is a better choice because its
descriptor dimension is  much smaller than overlapping $3/4$.

In conclusion, the CS-LBP feature based approach has the following descriptions:   $64 \times 128$
detection windows, $32\times32$ pixels block of four
$16\times16$ pixels cells; overlapping $1/2$ block (block spacing stride of $16$ pixels); the
Gaussian with $ \sigma =16$; $\ell_1$-\SQRTa block descriptor normalization; the descriptor length of
each detection window is $1334$ ($3 \times 7
\times 4 \times 16$ ); the detection window slides with a fixed step size of 8 pixels and a fixed scale factor of 1.09
in the 3D position and scale space.

\section{Pedestrian detection using pyramid CS-LBP/LTP features }
\label{sec:PLBPApproach}

Motivated by the image pyramid representation in~\cite{pyramidRepresentECCV06} and the HOG feature~\cite{C1},
Bosch et al.~\cite{spatialPyramidkernel} proposed the PHOG descriptor, which consists of a pyramid
of histograms of orientation gradients, to represent an image by its local shape and the spatial
layout of the shape. Experiments showed that the PHOG feature together with the histogram intersection
kernel can bring significant performance to object classification and recognition. Maji et
al.~\cite{Multi-levelHOG} introduced the PHOG feature into pedestrian detection and achieved the
current state-of-the-art on pedestrian detection. In this section, we propose the pyramid
CS-LBP$/$LTP features based pedestrian detection approach.

\subsection{The pyramid CS-LBP$/$LTP features}
\label{sec:PLBPApproach:FeaExtraction}
\begin{figure*}[t]
      \begin{center}
         \includegraphics[width=0.88\textwidth]{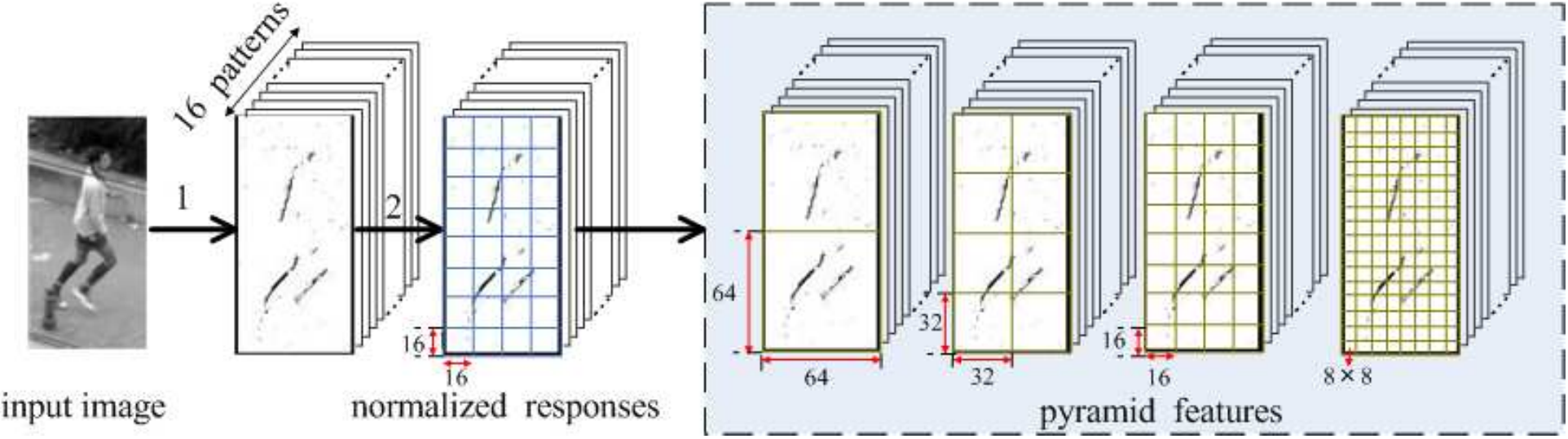}
      \end{center}
      \caption{The first three steps of computing the pyramid CS-LBP feature. (1) Edge energy
      responses corresponding to each CS-LBP pattern of the input image are computed. (2) The
      responses are $\ell_1$ normalized over all layers in each non overlapping $16\times16$ cells independently so that the normalized gradient values in each cell sum to unity. (3) The features at each level is extracted by concatenating the histograms, which are constructed by summing up the normalized response within each cell at the level. The cell size at level 1, 2, 3 and 4 are $64\times64$, $32\times32$, $16\times16$ and $8\times8$ respectively.}
      \label{multifeature}
   \end{figure*}

\begin{figure*}[t]
      \begin{center}
         \includegraphics[width=0.88\textwidth]{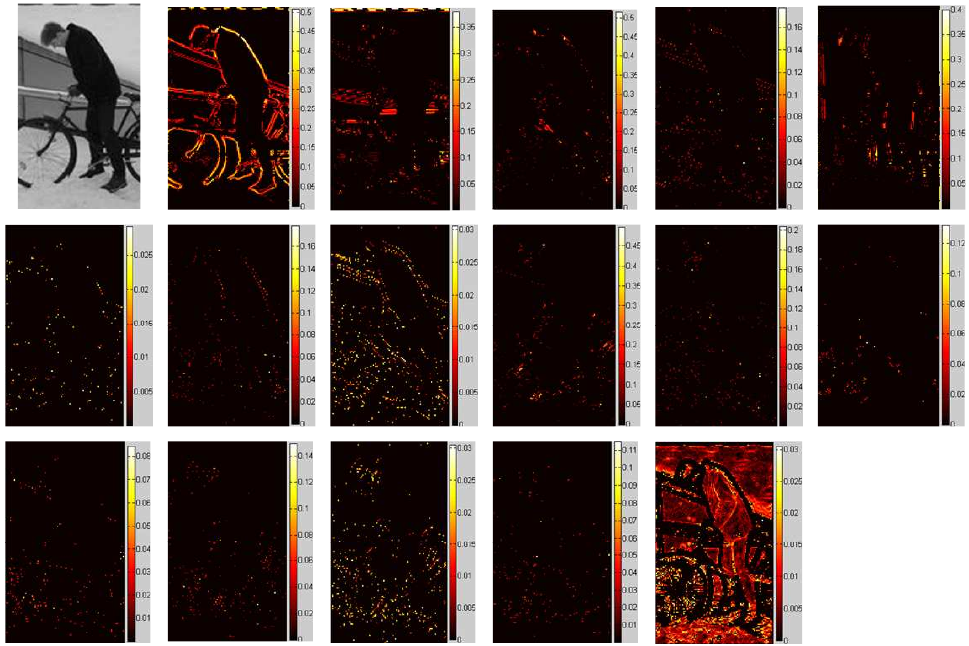}
      \end{center}
      \caption{Edge energy responses of an example image. The first image is the input image and the rests are its 16 layers of edge energy responses corresponding to each CS-LBP pattern. In order to show the response images more clearly, the response images are plotted out in color format by indexing the values to hot colormap. On the right of every response iamges shows the corresponding colorbar.}
      \label{edgeResponse}
   \end{figure*}

Because the LTP patterns can be divided into two separate channels of LBP patterns, we only
illustrate the computation of the pyramid CS-LBP features. Our features of a $64\times128$ detection
window are computed as follows ( Fig.~\ref{multifeature} shows the first three steps of computing
the features):
\begin{enumerate}
\item
We compute the CS-LBP value and the gradient magnitude of each  pixel of the input grayscale image
(detection window). The CS-LBP value is computed as \ref{e2} with $t=0.022$. Then we obtain
16 layers of gradient magnitude images corresponding
to each CS-LBP pattern. We call them edge energy responses of the input image.
Fig.~\ref{edgeResponse} shows the 16 layers of edge energy responses of the example image from INRIA
dataset. We can see that the first layer mainly captures the contours, the 16th layer mainly
captures the detailed textures or cluttered background, the rests capture spacial edges or textures.
The responses in the first layer is much bigger than those in the 16th layer. That is because
contours are more important than detailed textures to detect a pedestrian. Sometimes the detailed
textures (e.g., textures on the clothes of pedestrians) are harmful to pedestrian detection.
\item
Each layer of the response image is $\ell_1$ normalized in non overlapping cells of fixed size $y_n
\times x_n$ ($y_n=16$, $x_n=16$) so that the normalized gradient values in each cell sum to unity.
\item
At each level $l \in \{1,2,...L\}$, the response image is divided into non overlapping cells of
size $ y_l \times x_l $, and a histogram with 16 bins is constructed by summing up normalized response
within the cell. In our case, $L=4$, $y_1=x_1=64$, $y_2=x_2=32$, $y_3=x_3=16$, $y_4=x_4=8$. So we obtain 2, 8, 32, and 128 histograms at level $l=1$, 2, 3 and 4 respectively.
\item
The histograms of each level is normalized to sum to unity. This normalization ensures
that the edge or texture rich images are not weighted more strongly than others.
\item
The features at a level $l$ are weighted by a factor $w_l$ ($w_1=1$, $w_2=2$, $w_3=4$, $w_4=9$), and the features at all the levels are
concatenated to form a vector of dimension $2,720$, which is called pyramid CS-LBP features.
\end{enumerate}

The precess of computing pyramid uniform CS-LBP features is almost same as pyramid CS-LBP. The only
difference lies in the first step. In the first step, the edge energy responses corresponding to the 8 different uniform
patterns are count into 8 different layers and the edge energy responses corresponding to all the 8 non-uniform
patterns are count into one layer. So we obtain 9 layers of edge energy responses of the input image.

\subsection{Pedestrian detection based on pyramid CS-LBP$/$LTP features }

The first major component of our approach is feature extraction.
We perform the graylevel normalization of the input image so that the input image have the graylevel ranged from 0 to 1. Then the detection window slides on the input images in all positions and scales, with a fixed step size $8\times8$ and a fixed scale
factor $1.09$. We follow the steps in Sec.~\ref{sec:PLBPApproach:FeaExtraction} to compute the
pyramid CS-LBP$/$LTP features of each $64\times128$ detection window.

The second major component of our approach is the classifier. We use IKSVMs~\cite{Multi-levelHOG} as the classifier. The histogram intersection
kernel,
  \begin{equation}
   \label{eHIK}
       k_{\rm{HI}}(h_a,h_b) = \sum_{i=1}^n {\rm{min}}(h_{a}(i),h_{b}(i))
  \end{equation}
 was original proposed by Swain and Ballard~\cite{ColorIndexing} for color-based object recognition and has been shown to be a suitable measurement of similarity between histogram $h_a$ and $h_b$ ( $n$ is the length of the histogram). It is further shown to be positive definite~\cite{HIKdefinite} and can be used as a kernel for classification using SVMs. Compared to linear SVMs, histogram intersection kernel involves great computational expense.  Maji {et al.}~\cite{Multi-levelHOG,Maji_max-marginadditive} approximated the histogram intersection kernel for faster execution. Their experiments showed that the approximate IKSVMs consistently outperform linear SVMs at a modest increase in running time.

 The third major component of our approach is the merging of the multiple overlapping detections using non maximal suppression~\cite{HOGphdthesis}. After merging, detections with bounding boxes and confidence scores are obtained.

%-------------------------------------------------------------------------
\section{Experiments}
\label{sec:exp}

\subsection{Experiment setup}
\label{sec:exp:setup}

\textbf{Datasets.} We perform the experiments on INRIA human dataset \cite{C1}, which is one of the most
popular publicly available datasets. The datasets consist  of a training set and a test set.  The training set contains $1,208$ images of size $96 \times 160$ pixels (a margin of $16$ pixels around each side) of human samples ($2,416$ mirrored samples)
and $1,218$ pedestrian-free images. The test set contains $288$ images with $589$ human samples and $453$ human free images. Besides, in the test set, there is a fold contains 566 human samples ($1,132$ mirrored samples) of size $70\times134$ (a margin of 3 pixels around each side), which were cropped out from the 288 positive test images. All the human samples are cropped from a varied set of personal photos and vary in pose, clothing, illumination, background and partial occlusions, what make the dataset is very challenge.

\textbf{Methodology.}
\textsl{Per-window} performance is accepted as the methodology for evaluating pedestrian detectors
by most researchers. But this evaluating methodology is flawed. As pointed out in~\cite{dollarCVPR09peds},
\textsl{per-window} performance can fail to predicate \textsl{per-image} performance. There may be
at least two reasons: first, \textsl{per-window} evaluation does not measure errors caused by
detections at incorrect scales or positions or arising from false detections on body parts, nor does
it take into account the effect of non maximal suppression. Second, the \textsl{per-window} scheme uses cropped
positives and uncropped negatives for training and testing: classifiers may exploit window boundary
effects as discriminative features leading to good \textsl{per-window} but poor \textsl{per-image}
performance. In this paper, we use \textsl{per-image} performance, plotting detection rate versus
false positives per-image (FPPI).

We select the $2,416$ mirrored human samples from the training set as positive training examples.  A
fixed set of $12,180$ patches sampled randomly from $1,218$ pedestrian-free training images as initial
negative set. As in~\cite{C1}, a preliminary detector is trained and the $1,218$ negative
training images are searched exhaustively for false positives (`hard examples'). The final classifier is then trained
using the augmented set (initial $12,180$ + hard examples). The SVM tool we used is
LIBSVM~\cite{libsvm} and the fast intersection kernel SVMs tool we used is proposed by Maji et
al.~\cite{Multi-levelHOG}.

We detect pedestrians on each test images (both positive and negative) in all positions and scale
with a step size $8\times8$ and a scale factor $1.09$. Multiscale and nearby detections are merged
using non maximal suppression and a list of detected bounding boxes are given out. Evaluation on the
list of detected bounding box is done using the PASCAL criterion which counts a detection to be
correct if the overlap of the detected bounding box and ground truth bounding box is greater than
$0.5$.

\subsection{Detection results}

\begin{figure}[t] 
     \begin{center} 
         \includegraphics[width=0.42\textwidth]{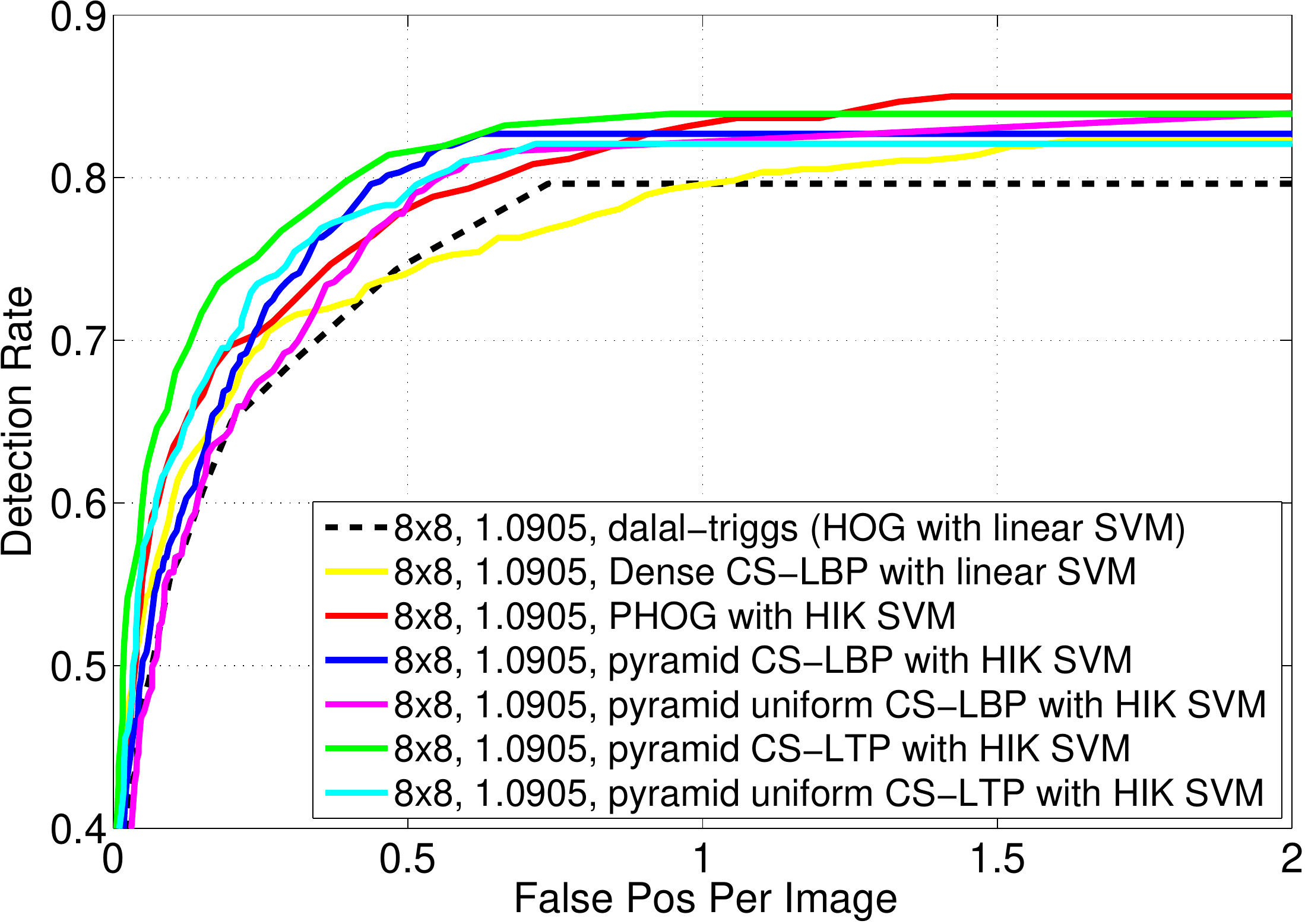} 
     \end{center}
     \caption{Detection rate versus false positive per-image (FPPI) curves for detectors based on
     the pyramid CS-LBP$/$LTP features using IKSVM classifier, the pyramid uniform CS-LBP$/$LTP
     features using IKSVM classifier, the PHOG feature using IKSVM classifier, the HOG feature using
     linear SVM classifier and the dense CS-LBP feature with linear SVM classifier. $8\times8$ is
     the step size and $1.0905$ is the scale factor of the sliding detection window.}
     \label{performance} \end{figure}

In this section, we study the performance of our dense CS-LBP feature based approach and the pyramid
CS-LBP$/$LTP features based approach by comparing with the HOG feature and the PHOG feature based
approaches. We obtain the HOG and the PHOG detectors from their authors, and all the parameters of
the PHOG (such as the $\ell_1$ normalization cell size, the level number and cell size in each
level) are same as our pyramid features. The results are shown in Fig.~\ref{performance}. The
performance of pyramid CS-LTP based detector performs best, with detection rate over $80\%$ at 0.5
FPPI. Then followed by the pyramid uniform CS-LTP based detector, which is slightly better than the
PHOG based detector. The pyramid CS-LBP based detector performs almost as good as the PHOG. Though
the pyramid uniform CS-LBP based detector performs slightly worse than PHOG basd detector, it
outperforms the HOG features with linear SVMs based detector proposed by Dalal and Triggs~\cite{C1}.
The performance of the dense CS-LBP feature with linear SVMs based detector is very close to the HOG
features with linear SVMs based detector. The results also show that the pyramid features with
HIKSVMs approach is more promising than the dense feature with linear SVMs approach.

%-------------------------------------------------------------------------
\subsection{Study on the features combined with the pyramid CS-LBP and PHOG}

In this experiment, our main aim is to find out whether the combination of our feature with the PHOG
feature can achieve better detection result or not. Feature Combination is a recent trend in
class-level object recognition in computer vision. One efficient method is to combine the kernels
corresponding to different features. The simplest method to combine several kernels is to average
them. Gehler and Nowozin~\cite{OnFeatureCombinition} pointed out that this simplest method is highly
competitive with multiple kernel learning (MKL)~\cite{MKL} method and the method based on boosting
approaches proposed in~\cite{OnFeatureCombinition}. Here, We simply average the two kernels
corresponding to the pyramid uniform CS-LBP feature and the PHOG feature as follows:
 \begin{equation}
 \label{multiclassifier}
 K_{\rm c}(v_1,v_2)= \tfrac{1}{2} [ K_1(v_1) + K_2(v_2) ],
 \end{equation}
where $K_1$ and $K_2$ are the IKSVMs classifiers pretrained using the pyramid uniform CS-LBP feature
and the PHOG feature respectively, $v_1$ and $v_2$ are the pyramid uniform CS-LBP feature and the
PHOG feature of a detection window respectively.

\begin{figure}[t]
      \begin{center}
         \includegraphics[width=0.42\textwidth]{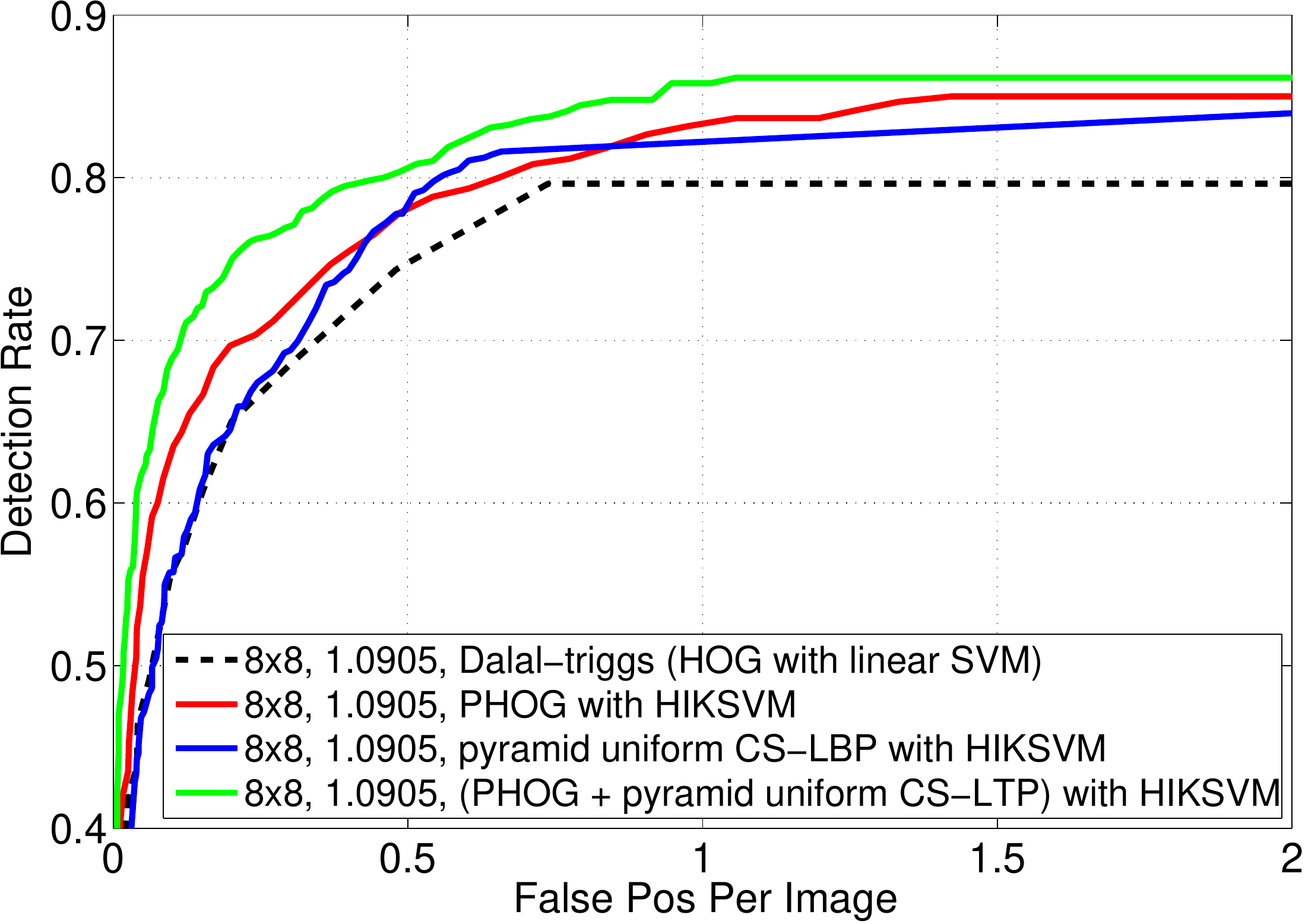}
      \end{center}
      \caption{Detection rate versus false positive per-image(FPPI) curves for detectors(using IKSVM classifier) based on the PHOG features, the uniform CS-LBP feature and the augmented features combined by the HOG and the pyramid uniform CS-LBP. The augmented feature can improve the detection accuracy significantly. $8\times8$ is the step size and $1.0905$ is the scale factor of the sliding detection window.
      }
      \label{combine-performance}
   \end{figure}

\begin{figure*}[!h]
      \begin{center}
         \includegraphics[width=0.998\textwidth]{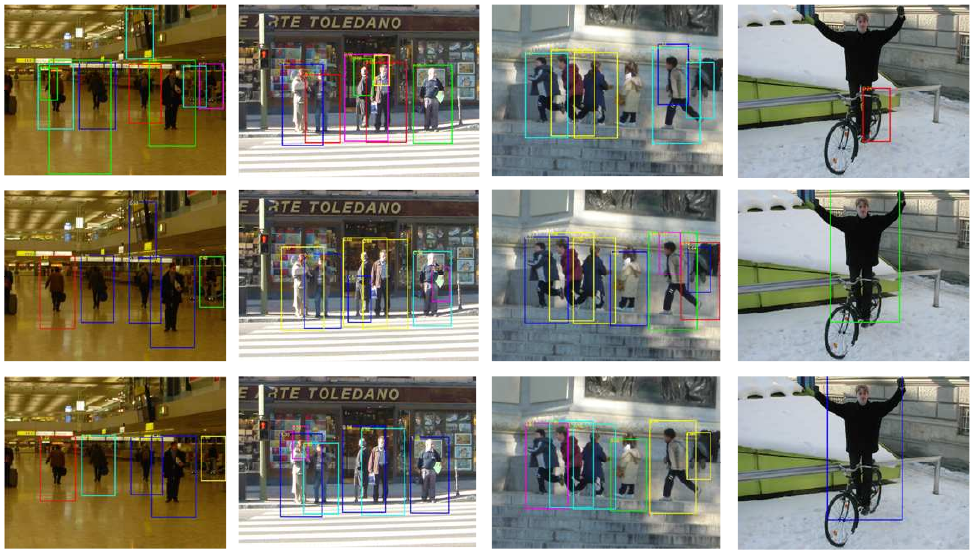}
      \end{center}
      \caption{Some examples of detections on test images for the detectors using PHOG, pyramid
      uniform CS-LBP and augmented features (combined with HOG and pyramid uniform CS-LBP). First row:
      detected by the PHOG based detector. Second row: detected by the pyramid uniform CS-LBP based
      detector. Third row: detected by the PHOG+pyramid uniform CS-LBP based detector.}
      \label{detectionExamples}
   \end{figure*}

Detection performance are shown In Fig.~\ref{combine-performance}. The detection rate versus FPPI
curves show that the feature combination can significantly improve the detection performance.
Compared to the PHOG, the detection rate raises about $6\%$ at 0.25 FPPI and raises about $1.5\%$ at
0.5 to 1 FPPI. Fig.~\ref{detectionExamples} shows pedestrian detection on some example test images.
The three rows show the bounding boxes detected by PHOG based detector, the pyramid uniform CS-LBP
based detector and the PHOG + pyramid uniform CS-LBP based detector, respectively. We can see that
the PHOG with pyramid uniform CS-LBP based detector performs best.

%===========================================================

\section{Conclusion}
\label{sec:conclusion}

We have presented the dense CS-LBP feature and the pyramid CS-LBP$/$LTP features for
pedestrian detection. Experimental results on the INRIA dataset show that the dense CS-LBP feature
based approach the pyramid CS-LTP features using the IKSVM classifier outperform the PHOG, and the
pyramid CS-LBP features perform as well as the HOG feature.  We have also show that combining the
pyramid CS-LBP with PHOG produces a significantly better detection performance on the INRIA
dataset.

There are many directions for further research. To make the conclusion more convincing, the
performance of the pyramid CS-LBP$/$LTP features based pedestrian detector needs  to be further
evaluated on other dataset, e.g., the Daimler Chrysler Pedestrian
Dataset~\cite{ExperimentalPedestrianClassification} and the Caltech Pedestrian
Dataset~\cite{dollarCVPR09peds}. Another further study is to compare the computational complexity of
the pyramid CS-LBP$/$LTP features with PHOG both theoretically and experimentally. Thirdly, it is
worthy studying how to combine our features with PHOG or other features more efficiently.  We are
also interested in implement the new feature in a boosting framework.

\bibliographystyle{ieee}
\bibliography{final}

\begin{IEEEbiography}
    %[{\includegraphics[width=1in,height=1.25in,clip,keepaspectratio]{yongbin.jpg}}]
    {Yongbin Zheng}
        is a Ph.D. student at the National University of Defense Technology, China and is currently
        visiting the Canberra Research Laboratory of NICTA (National ICT Australia) and the
        Australian National University. He received the B.E. degree and M.E. degree in Control
        Science from the National University of Defense Technology in 2004 and 2006.  His research
        interests include pattern recognition, image processing and machine learning.
\end{IEEEbiography}
\begin{IEEEbiography}
    % [{\includegraphics[width=1in,height=1.25in,clip,keepaspectratio]{shen}}]
    {Chunhua Shen}
        completed the Ph.D. degree from School of Computer Science,
        University of Adelaide,
        Australia in 2005; and the M.Phil. degree from
        Mathematical Sciences Institute, Australian National University,
        Australia in 2009. Since Oct. 2005, he has been
        working with the
        computer vision program, NICTA (National ICT Australia),
        Canberra Research Laboratory, where he is a senior research fellow
        and holds a continuing research position.
        His main research interests include
        statistical machine learning and its applications in computer vision and image processing.
\end{IEEEbiography}

\begin{IEEEbiography}
   %[{\includegraphics[width=1in,height=1.25in,clip,keepaspectratio]{Hartley.jpg}}]
    {Richard Hartley}
        received the degree from the University of Toronto in 1976 with a thesis on knot theory.
        He is currently with the computer vision group at the Australian National University 
        and also with NICTA (National ICT Australia), a government-funded research institute.
        He worked in this area for several years before joining the General Electric Research and 
        Development Center, where he developed a computer-aided electronic design system called 
        the Parsifal Silicon Compiler, described in his book
        ``Digit Serial Computation''.
        Around 1990, he developed an interest in computer vision, 
        and in 2000, he coauthored (with Andrew Zisserman) a book on multiple-view geometry.
        He has written papers on knot theory, geometric voting theory, computational geometry,
        computer-aided design, and computer vision.
        He holds 32 US patents. In 1991, he was awarded GE's Dushman Award.
        He is a fellow of the IEEE and a member of the IEEE Computer Society.
   \end{IEEEbiography}

\begin{IEEEbiography}
  %  [{\includegraphics[width=1in,height=1.25in,clip,keepaspectratio]{Xinsheng Huang.jpg}}]
    {Xinsheng Huang}
        is a professor at the College of Mechatronics Engineering and Automation,
        the National University of Defense Technology, China.
        His research interests include control science, inertial navigation and image processing.
\end{IEEEbiography}

\end{document}